# Myopic Value of Information in Influence Diagrams

Søren L. Dittmer
DINA Skejby
The Danish Agricultural Advisory Centre
Udkærsvej 15, Skejby
DK – 8200 Århus N, Denmark
E-mail: `dittmer@lr.dk`

Finn V. Jensen
Department of Computer Science
Aalborg University
Fredrik Bajers Vej 7E
DK – Aalborg Øst, Denmark
E-mail: `fvj@cs.auc.dk`

## Abstract

We present a method for calculation of myopic value of information in influence diagrams (Howard & Matheson, 1981) based on the strong junction tree framework (Jensen et al., 1994).

An influence diagram specifies a certain order of observations and decisions through its structure. This order is reflected in the corresponding junction trees by the order in which the nodes are marginalized. This order of marginalization can be changed by table expansion and use of control structures, and this facilitates for calculating the expected value of information for different information scenarios within the same junction tree. In effect, a strong junction tree with expanded tables may be used for calculating the value of information between several scenarios with different observation-decision order.

We compare our method to other methods for calculating the value of information in influence diagrams.

## 1 INTRODUCTION

Influence diagrams were introduced by Howard & Matheson (1981) as a formalism to model decision problems with uncertainty for a single decision maker.

An influence diagram can be considered a Bayesian network augmented with decision variables and a utility function. The decision variables, $D_1, \ldots, D_n$, in the influence diagrams are partially ordered and the chance variables are divided into information sets, $I_0$, $\ldots, I_n$. The information set $I_{i-1}$ is observed immediately before decision $D_i$ is made, and the information set $I_n$ consists of the chance variables that are observed later than the $n$'th decision is made, if ever.

Let $V_i$ be the set of variables preceding $D_i$, that is, $V_i$ contains the past relevant for $D_i$. The solution of a decision problem modeled by an influence diagram is a sequence of decisions that maximizes the expected utility. Shachter (1986) describes a method to solve an influence diagram without unfolding it into a decision tree; rather, the influence diagram is transformed through a series of node-removal and arc-reversal operations. Shenoy (1992) describes another approach to the problem of solving influence diagrams by conversion into valuation networks. This approach is slightly more efficient than that of (Shachter, 1986). (Shachter & Ndilikilikesha, 1993) and (Ndilikilikesha, 1994) modified the node-removal/arc-reversal algorithm and achieved a method that is equivalent to the algorithm presented in (Shenoy, 1992) with respect to computational efficiency.

Jensen et al. (1994) describes an efficient method for solving influence diagrams using strong junction trees. This is an extension to the junction trees used for computation in pure Bayesian decision analysis. It is on this framework we base the present work.

We are about to choose among a set of $k$ options. These options are packed into the decision node $D$. We have already received some information $e$, and now we can either choose among the options or we can look for more information. The 'looking for more information' is to consult some source which will provide the state of a chance variable. Let the chance variables in question be the set $\Gamma = \{A_1, \ldots, A_m\}$. We want to calculate what we can expect to gain from consulting the information source. For all the considerations in this paper we deal with the *myopic* value-of-information question: At any time, we can ask for the state of at most one of the variables in $\Gamma$.

As basis for the considerations we have $EU(D|e)$, the expected utilities for $D$ given the evidence $e$, and the decision $d$ of maximal expected utility is chosen. If $A_i \in \Gamma$ is observed to be in state $a$, then $EU(D|e, A_i = a)$ is the new basis. Now, before observing $A_i$ we have probabilities $P(A_i|e)$, and the expected utilities of the optimal action after having observed $A_i$ is

$$\mathrm{EUO}(A_i, D|e) = \sum_{A_i} P(A_i|e) \cdot \max_D \mathrm{EU}(D|e, A_i)$$

The *value of observing* $A_i$ is the difference

$$\mathrm{VOI}(A_i, D|e) = \mathrm{EUO}(A_i, D|e) - \max_{D_i} \mathrm{EU}(D|e)$$

Value of information is a core element in decision analysis, and a method for efficient calculation of myopic value of information in Bayesian networks (augmented with a utility function) is described by (Jensen & Jiangmin L., 1995). Also, (Heckerman et al., 1992) describes a method for calculating the utility-based myopic value of information.

Methods for computing the value of information in influence diagrams have been described by (Ezawa, 1994) based on the arc-reversal/node-removal methods. (Poh & Horvitz, 1996) approach a notion of qualitative value of information through graph-theoretic considerations yielding a partial order of the chance nodes in the model.

The value of information can be viewed as the difference in expected value between two models only differing in the observation-decision sequence in the influence diagram. We present a single-model framework for calculating the exact value of information of a chance node.

For the considerations in this paper, the network is of considerable size so that a propagation in the network is a heavy (but feasible) task. This means that the methods presented shall be evaluated in the light of their propagation demand.

## 2   SIMPLE SCENARIOS

We shall first describe a couple of simple scenarios which have efficient solutions. The first scenario is standard and has been treated more detailed by (Jensen, 1996).

### 2.1   ONE NON-INTERVENING DECISION

There is one decision node $D$ which has no impact on any of the chance nodes in the model. The utility function $U$ is a function of $D$ and the chance variable $H$ which may actually be a set of variables (see Figure 1).

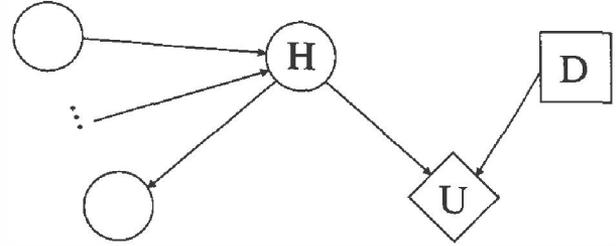

Figure 1: The scenario with one non-intervening decision node.

For this scenario we have

$$\begin{aligned}
\mathrm{VOI}&(A_i, D|e) \\
&= \sum_{A_i} P(A_i|e) \cdot \max\left(\sum_H P(H|A_i, e) \cdot U(D, H)\right) \\
&\quad - \max_D \left(\sum_H P(H|e) \cdot U(D, H)\right)
\end{aligned}$$

For the calculation of $\mathrm{VOI}(A_i, D|e)$ we need $P(H|A_i, e)$ for all variables $A_i$ in $\Gamma$. These conditional probabilities can be achieved through entering and propagating each state of $A_i$. Using Bayes' rule, the requirement is transformed to a need for $P(A_i|H, e)$ for all $A_i$ in $\Gamma$. They can be achieved all by entering and propagating the states of $H$. So, the number of propagations necessary for solving the value-of-information task for this scenario is the minimum of the number of states of $H$ and the sum of the states of the variables in $\Gamma$.

### 2.2   THE NUMBER OF $H$ IS LARGE

The assumptions in Section 2.1 are very crude and we would like to relax them. Often $D$ has an impact on $H$ and in that case we will need $P(H|D)$. Also, the number of states of $H$ as well as the sum of all states of $\Gamma$ may be very large ($H$ may be a large set of variables), and we will look for methods requiring less propagations (see Figure 2).

The following method reduces the number of propagations to the number of states in $D$. The method is a modification of a trick by Cooper (1988). The utility function is transformed to a normalized utility NU through a linear transformation such that $0 \le \mathrm{NU} \le 1$. NU is represented in the influence diagram by a binary node $NU$ with the argument variables $H$ (which might include $D$) as parents and with $P(NU = y|H) = \mathrm{NU}(H)$ (see Figure 3).

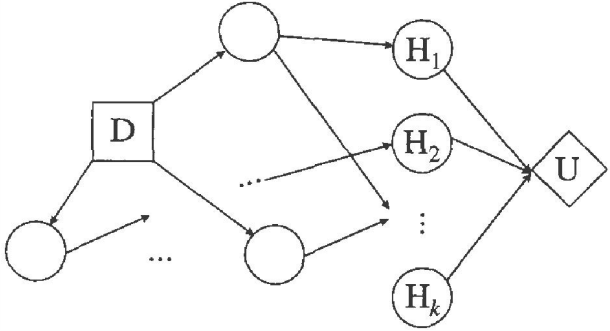

Figure 2: A scenario where the method of Section 2.1 is inadequate.

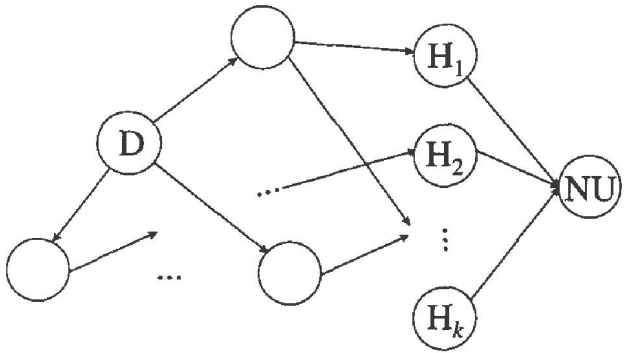

Figure 3: The Cooper transformation of the scenario in Figure 2.

The normalized value of information is defined as

$\text{NVOI}(A_i, D|e)$
$= \sum_{A_i} P(A_i|e) \cdot \max_D \left( \sum_H P(H|A_i, D, e) \cdot \text{NU}(D) \right)$
$\quad - \max_D \left( \sum_H P(H|D, e) \cdot \text{NU}(H) \right)$

and VOI can be calculated from NVOI by the inverse transformation.

The expected normalized utility of a decision $d$, given the evidence $e$ can be calculated as

$\text{ENU} = \sum_H \text{NU}(H) \cdot P(H|d, e)$
$\quad = \sum_H P(NU = y|H) \cdot P(H|d, e)$
$\quad = \sum_H P(NU = y, H|d, e)$
$\quad = P(NU = y|d, e)$

Using Bayes' rule and giving $D$ the even distribution, $\text{ENU}(D|e)$ can be calculated by entering and propagating $NU = y$.

Now, let $A$ be a variable in $\Gamma$. Assume that $A$ is observed to be in the state $a$. Then we have

$\text{ENU}(D|a, e)$
$\quad = P(NU = y|D, a, e)$
$\quad = P(NU = y|D, e) \cdot \dfrac{P(a|NU = y, D, e)}{P(a|D, e)}$
$\quad = \text{ENU}(D|e) \cdot \dfrac{P(a|NU = y, D, e)}{P(a|D, e)}$

and the expected normalized utility after observing $A$ is $\sum_A (\max_D \text{ENU}(D|A, e)) \cdot P(A|D, e)$.

The required probabilities $P(A|NU = y, D, e)$ and $P(A|D, e)$ can be achieved by entering and propagating the states of $D$ in a network conditioned on $e$ and in one conditioned on $(e, NU = y)$. Hence, the number of propagations required for this calculation is twice the number of states in $D$, that is, with $2k$ propagations we can calculate the value of observation for all variables. It should be noted that there were no structural assumptions for this result.

In most cases the information $e$ as well as the variables which may be observed prior to $D$ are not descendants of $D$. In these cases $P(A|D, e) = P(A|e)$ and the method only requires $k$ propagations.

## 3 A SEQUENCE OF DECISIONS

The next scenario to consider is the following: We have a sequence of decisions and observations $I_0, D_1, I_1, \ldots, D_n, I_n$ where each $I_i$ is a set of chance variables ($I_n$ is the set of variables which are never observed). The variables are structured in an influence diagram (see Figure 4 for an example). We are in the middle of this sequence, we have observed $I_{i-1}$ and are about to decide on $D_i$ but we have a further option of observing *one* variable of the set $\Gamma$. Let $\text{VOI}(X, Di, , j|Vi)$ (where $i < j$) denote the difference in maximal expected utility for $D_i$ between observing chance node $X$ immediately before deciding on $D_i$ and immediately before deciding on $D_j$. That is $\text{VOI}(X, Di, j|Vi)$ denotes the difference between having $X$ in $I_{i-1}$ and in $I_{j-1}$ at the time of deciding on $D_i$.

The standard dynamic programming technique for solving an influence diagram is to perform a sequence of marginalizations in reverse order (Shenoy, 1992; Shachter & Peot, 1992). Chance nodes are marginalized through a summation and decision nodes are maximized. Since summation and maximization do not commute, the order of marginalization is impor-tant and it is performed in the following order: First marginalize $I_n$ (in any order), then $D_n$, then $I_{n-1}$ (in any order), etc. When $I_i$ has been marginalized, we have a representation of the expected utility of the various options of $D_i$ given the past.

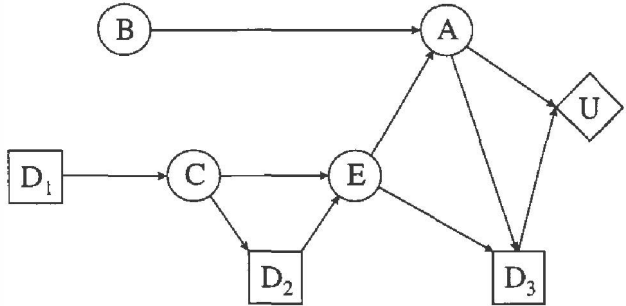

Figure 4: An influence diagram with the observation-decision sequence $D_1$, $C$, $D_2$, $\{A, E\}$, $D_3$, $B$. Note that $A$ and $E$ may be observed in mutually arbitrary order but both *will* be observed.

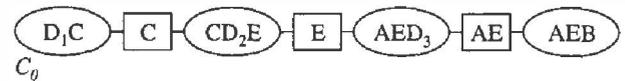

Figure 5: A strong junction tree for the influence diagram in Figure 4. The strong root is the clique $C_0$ at the far left.

It is tempting to use this technique to condense the future into a utility function over a subset of the currently unknown variables and the decision node $D_i$ and to use this condensed future for the calculation of value of information. However, the condensed future contains max-expected-utility decisions, and observing a variable from $\Gamma$ may affect these decisions. This can be avoided by assuming that the future is independent of $\Gamma$ given $D_i$ (and the past). Such an assumption will rarely hold, and instead we will introduce a technique which does not have that kind of assumption.

In (Jensen et al., 1994) the junction tree technique is used to solve influence diagrams. A so-called *strong junction tree* is constructed with a so-called *strong root*. This means that there is a clique $C_0$ such that when a collect-operation to $C_0$ is performed, then all marginalizations can be performed in the proper order (see Figure 5). Note that the strong junction tree in itself does not ensure that marginalizations are performed in a proper order. When marginalizing in a clique we need a *control structure* giving the order of marginalizations. The "proper order" need not be the reversed temporal order. It is sufficient that each variable is eliminated in reverse temporal order with respect to its Markov blanket. The Markov blanket of a node $X$ is the minimal set of nodes covering $X$ from influence from other nodes, that is, the Markov blanket for node $X$ consists of $X$'s parents, children, and children's parents.

In Figures 4 and 5, $B$ is not observed (or rather: $B$ is not observed until after the last decision is made). Now, assume that before deciding on $D_1$, we observe the chance variable $B$. The model for this observation-decision sequence is shown in Figure 6.

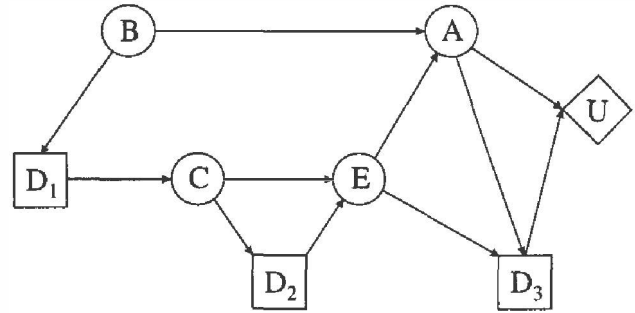

Figure 6: An influence diagram with the observation-decision sequence $B$, $D_1$, $C$, $D_2$, $\{A, E\}$, $D_3$.

The difference in expected utility when solving the two influence diagrams is $\text{VOI}(B, D_1, \infty | V_i)$, that is, the value of observing $B$ before $D_1$ rather than never observing $B$. The difference between the two scenarios can be seen on the strong junction trees in Figures 7a and 7b.

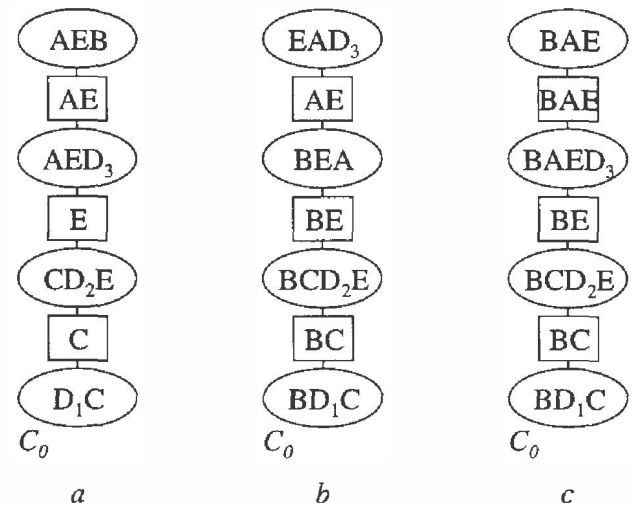

Figure 7: Strong junction trees for the two scenarios of Figures 4 and 6, and a junction tree adequate for both scenarios.

It is possible to construct a junction tree capable of solving both scenarios and in effect calculate the value of information between the two information scenarios. The crucial thing about a strong junction tree is that it allows marginalization in a proper (reverse) temporal order and this can be done for both temporal orders in the strong junction tree shown in Figure 7c. This strong junction tree is obtained from the junction tree in Figure 7a by adding $B$ to the cliques down to $(D_1, C)$.

This observation can be used in general: To obtain a strong junction tree with strong root $C_0$ for calculating VOI$(A, D_i, j|V_i)$, construct a strong junction tree for the scenario with $A$ in $I_{j-1}$. Then $C_0$ imposes a (partial) order $<$ for the cliques, such that $C < C'$ if and only if $C$ is on the path from $C'$ to $C_0$. Identify the cliques $C_i$ and $C_A$. $C_i$ is the clique closest to the $C_0$ containing $D_i$, and $C_A$ is the clique closest to $C_0$ containing $D_i$. Let $C_{iA}$ be the "greatest lower bound" of $C_i$ and $C_A$. That is, $C_{iA}$ is the clique furthest away from $C_0$ such that $C_{iA} < C_i$ and $C_{iA} < C_A$ (when the temporal order is strict, then $C_{iA} = C_i$). Finally, extend all cliques on the path between $C_{iA}$ and $C_A$ with the variable $A$.

As mentioned earlier, a control structure is associated with the (strong) junction tree. This structure handles the order of marginalization, and therefore we can use the expanded junction tree (and the associated control structure) in Figure 7c to marginalize $B$ from any clique of our choise. After $B$ has been marginalized from a clique, the table space reserved for $B$ in cliques closer to the strong root is obsolete. Clever use of the control structures will prevent calculations to take place in the remaining table expansions, and the number of table operations in the remaining subtree equals that of an ordinary strong junction tree.

### 3.1 NON-STRICT TEMPORAL ORDERS

As mentioned previously, a proper elimination order of an influence diagram is an order where the elimination order of each node and its Markov blanket is a reverse temporal order. This means that although the influence diagram in the offset requires a linear temporal order of the decisions, then the actual diagram may disclose temporal independencies which can be exploited when solving it.

The influence diagram in Figure 8 has a temporal order of the decision nodes with increasing index. However, when $f$ has been observed, then $D_3$ can be decided at any time independently of the observations and decisions on $e$, $g$, $D_2$, and $D_4$. This is also reflected in the strong junction tree in Figure 9a where the branch containing $D_3$ can be marginalized independently of the other branches.

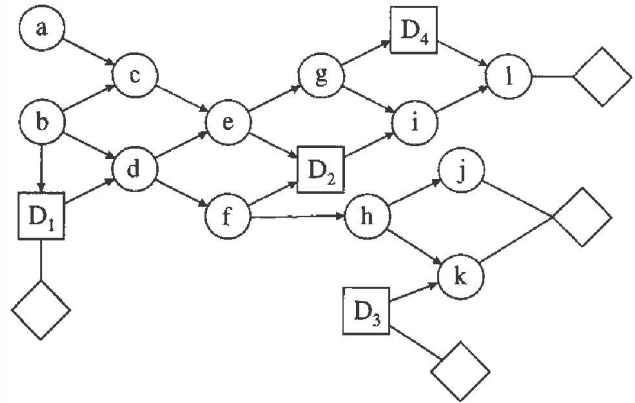

Figure 8: An influence diagram with temporal order from left to right (no-forgetting arcs are not included). It discloses temporal independence between $D_3$ and $\{D_2, D_4\}$. (From (Jensen et al., 1994))

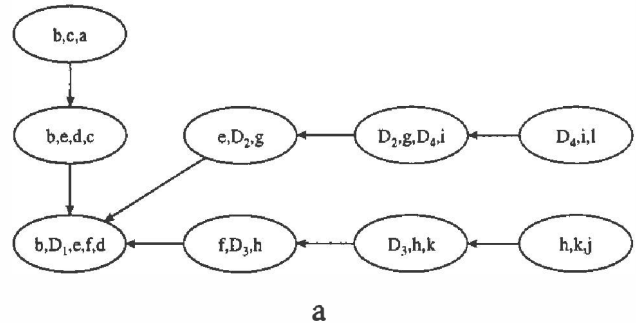

a

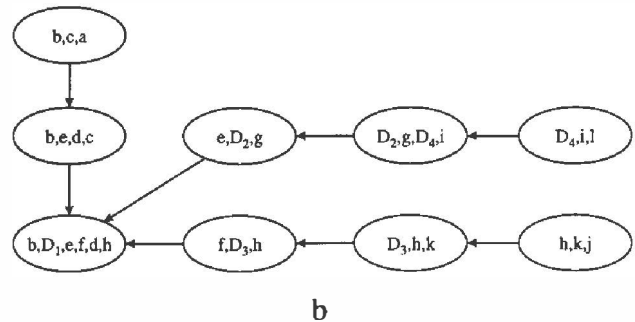

b

Figure 9: Strong junction trees (derived from Figure 8 illustrating the difference between never observing $h$ (a) and observing $h$ immediately before decision $D_2$ (b) when decisions are not strictly ordered.

The value of information technique is illustrated on the influence diagram in Figure 8 through Figures 9a and 9b. The strong junction tree in Figure 9a can also be used to solve an influence diagram with $h$ observed before deciding on $D_3$. The difference between the two scenarios is reflected in the control structure for the collect operation rather than in the junction tree. A strong junction tree also being able to handle the situation where $h$ is observed before deciding on $D_2$ is shown in Figure 9b.

### 3.2 NOTATION

In Figure 10, we present an extended version of the influence diagram from (Jensen et al., 1994). The original influence diagram notation has been extended with triangular nodes, observation nodes. An observation node designates that the chance node associated with it will be observed within some interval of information sets.

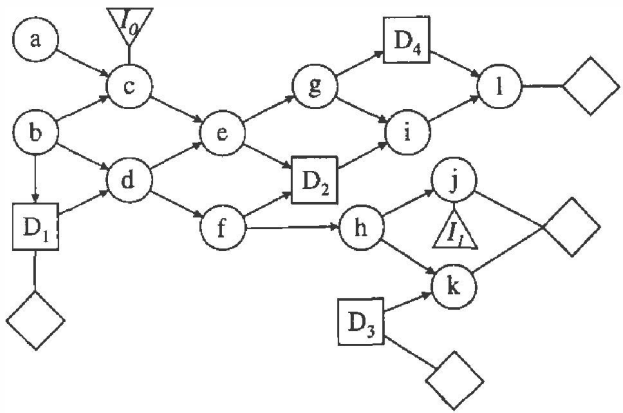

Figure 10: Influence diagram from (Jensen et al., 1994) with extended notation.

Though there may not be any computational difficulties associated with observing variables at an earlier time than modeled, there may be some conceptual problems. It does not make sense to observe, say, the state of a fungus attack on your crop in May before deciding whether or not to apply fungicide in April. In other words: We cannot observe a variable prior to making a decision that influences it.

Hence, a variable is modeled in the influence diagram as belonging to the last information set possible, and the observation node is associated with a "lower boundary" for the observation. For node $c$ in Figure 10 the lower boundary is $I_0$, yielding the observation interval to be $[I_0; I_4]$ whereas the lower boundary for node $j$ is $I_1$ and hence the observation interval for $j$ is $[I_1; I_4]$. If associated with an observation node, node $g$ would have the observation interval $[I_1; I_3]$.

### 3.3 ALTERNATIVE METHODS

There are other methods for calculating the value of information in influence diagrams. These can be separated into *multiple-model* methods and *single-model* methods.

The value of information in influence diagrams can be viewed as the difference in expected utility between a set of influence diagrams each implementing a specific scenario of the desired observation-decision sequences. In that view Ezawa (1994) creates and solves multiple models for calculating the value of information in influence diagrams. However, as the construction of strong junction trees is a complex task it is preferable to reduce the number of different junction trees. Also, to cover all desired observation-decision sequences the decision analyst may be facing a considerable task in constructing the needed influence diagrams.

Instead, the different decision models in Figure 4 and Figure 6 can be combined into a single influence diagram which gives us the power to calculate whether or not to observe $B$. Such a model is shown in Figure 11.

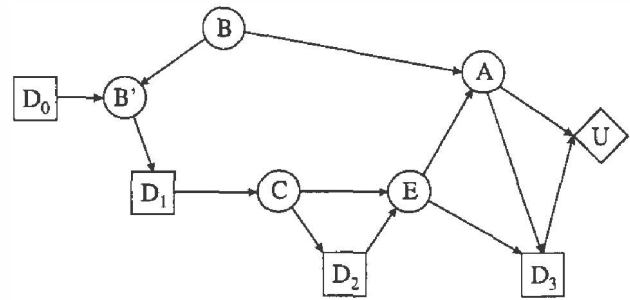

Figure 11: General model capable of handling the scenarios of Figures 4 and 6.

The resulting model consists of the original model without observation on $B$ (from Figure 4) with an additional two nodes; a decision node, $D_0$ and the chance node $B'$.

$D_0$ will consist of the decisions $(B)$ and $(\neg B)$ and the observed node, $B'$ will have the same states as its unobserved counterpart, $B$, plus an additional state, $(No\ observation)$. If the optimal decision, $d_0$, is $(B)$, then $B'$ is observed and set to the true state of $B$; if the optimal decision is $(\neg B)$, $B'$ is set to $(No\ observation)$. The probability table for $B$ is equal to the one specified in Figures 4 and 6 and the behavior of $B'$ is specified

as

$$B' = No\ observation \text{ for } d_0 = (\neg B)$$
$$B' = B \text{ otherwise}$$

This type of modeling cannot be called neither simple nor intuitive. Furthermore, as can be seen from Figure 12, the junction tree for the general model in Figure 11 is larger than the junction tree produced by expansion (Figure 7c).

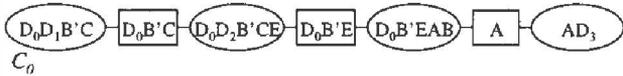

Figure 12: Strong junction tree constructed from the general model of Figure 11.

It is also worth noting that the model in Figure 11 and its corresponding junction tree in Figure 12 are made for the case where $B$ is either unobserved or observed before $D_1$. The junction tree in Figure 7c is capable of calculating the expected utility for the decision problem with $B$ belonging to any information set. Should the model in Figure 11 be extended to the same flexibility, we are facing a larger and considerably less intuitive model with little resemblance to the original decision problem.

## 4 CONCLUSION

For specific influence diagrams, such as scenarios with non-intervening decisions, we have presented a simple method for calculating the value of information. This method is simple in construction and cheap in terms of time and space requirements, but is restricted in the structure of the influence diagram. It is based on methods developed by (Cooper, 1988) and (Jensen & Jiangmin L., 1995). For certain, well-defined tasks there may be advantages in using this method but in the general case we propose to use the method presented for influence diagrams with sequences of intervening decisions.

In strong junction trees constructed for decision problems formulated as general influence diagrams we are able to calculate the value of information for a given chance node, that is, the gain in expected utility from observing variable $X$ before making a decision $D_i$. In other words, we can calculate the difference in expected utility between models that differ in observation-decision sequence, using the same junction tree structure with only a number of tables expanded but not recalculated. We find this method far more intuitive than modeling all possible outcomes in a general influence diagram as the structure of the model will not change even when chance nodes (within limits) are observed prior to the latest possible observation time. Also, modeling observations as intervening decisions may seem unappealing to decision analysts. In addition to this, we experienced that the junction trees produced from the general models are larger than those produced by table expansion.

Using our method is not for free as in its worst case (modeling a chance node as never observed and observing it before the first decision $D_1$) all tables in the junction tree will be expanded (assuming that the decisions are strictly ordered). This means that with $\alpha$ states in the node in question, the resulting junction tree will be almost $\alpha$ times larger than the original junction tree. This corresponds to performing $\alpha$ propagations in the strong junction tree and the gain is therefore minimal.

However, the method presented will only expand the tables needed, that is, only part of the junction tree becomes larger (by a factor of $\alpha$) which consequently reduces the number of operations performed during a propagation. Also, clever use of the control structures associated with the strong junction tree will prevent excess operations in the expanded tables after marginalization of the node in question. Still, if for example $\Gamma$ is very large and if all $A \in \Gamma$ are placed in $I_n$, we may very well face an intractable problem as we expand the cliques beyond the capacity of computers. Topics for further research include the possibility for utilizing independence assumptions in order to further reduce complexity.

## References

[Cooper, 1988] Gregory F. Cooper. A method for using belief networks as influence diagrams. In *Fourth Workshop on Artificial Intelligence*, pages 55 – 63, University of Minnesota, Minneapolis, 1988.

[Ezawa, 1994] Kazuo J. Ezawa. Value of evidence on influence diagrams. In R. L. de Mantaras and D. Poole, editors, *Uncertainty in Artificial Intelligence*, pages 212–220, San Francisco, California, July 1994. Morgan Kaufman.

[Heckerman et al., 1992] David E. Heckerman, Eric J. Horvitz, and Bharat N. Nathwani. Towards Normative Expert Systems: Part I. The Pathfinder Project. *Methods of Information in Medicine*, 31:90 – 105, 1992.

[Howard & Matheson, 1981] R. A. Howard and J. E. Matheson. Influence diagrams. In R. A. Howard and J. E. Matheson, editors, *Readings on the principles and applications of decision analysis*, volume 2, pages 719 – 762. Strategic Decisions Group, Menlo Park, CA, 1981.

[Jensen & Jiangmin L., 1995] Finn Verner Jensen and Jiangmin L. drHugin: A system for hypothesis driven data request. In *Probabilistic Reasoning and Bayesian Belief Networks*, pages 109 – 124. Alfred Waller, Ltd., London, 1995.

[Jensen *et al.*, 1994] Frank Jensen, Finn V. Jensen, and Søren L. Dittmer. From influence diagrams to junction trees. In R. L. de Mantaras and D. Poole, editors, *Proceedings of the Tenth Conference on Uncertainty in Artificial Intelligence*, pages 367 – 373, San Francisco, California, July 1994. Morgan Kaufmann.

[Jensen, 1996] Finn Verner Jensen. *An introduction to Bayesian Networks*. UCL Press, London, 1996.

[Ndilikilikesha, 1994] Pierre Ndilikilikesha. Potential influence diagrams. *International Journal of Approximate Reasoning*, 10(3), 1994.

[Poh & Horvitz, 1996] Kim Leng Poh and Eric Horvitz. A Graph-Theoretic Analysis of Information Value. In Eric Horvitz and Finn Jensen, editors, *Proceedings of the Twelfth Conference on Uncertainty in Artificial Intelligence*, pages 427 – 435, San Francisco, CA, 1996. Morgan Kaufman.

[Shachter & Ndilikilikesha, 1993] Ross D. Shachter and Pierre Ndilikilikesha. Using potential influence diagrams for probabilistic inference methods. In *Proceedings of the Ninth Conference on Uncertainty in Artificial Intelligence*, pages 383 – 390, San Mateo, CA, 1993. Morgan Kaufmann.

[Shachter & Peot, 1992] Ross D. Shachter and Mark A. Peot. Decision making using probabilistic inference method. In Didier Dubois, Michael P. Wellman, Bruce D'Ambrosio, and Phillipe Smets, editors, *Uncertainty in Artificial Intelligence, 8*, pages 276 – 283. Morgan Kaufmann, 1992.

[Shachter, 1986] Ross D. Shachter. Evaluating influence diagrams. *Operations Research*, 34(6):871 – 882, November-December 1986.

[Shenoy, 1992] Prakash P. Shenoy. Valuation-based systems for bayesian decision analysis. *Operations Research*, 40(3):463 – 484, 1992.